\documentclass[conference]{IEEEtran}
\IEEEoverridecommandlockouts
\usepackage{amsmath,amsfonts,extpfeil}
\usepackage{algorithmic}
\usepackage{algorithm}
\usepackage{array}
\usepackage[caption=false,font=normalsize,labelfont=rm,textfont=rm]{subfig}
\usepackage{textcomp}
\usepackage{stfloats}
\usepackage{url}
\usepackage{verbatim}
\usepackage{graphicx}
\usepackage{cite}
\usepackage{lineno,hyperref}
\usepackage{graphicx}
\usepackage{epstopdf}
\usepackage{amssymb}
\usepackage{amsmath}
\usepackage{enumitem}
\usepackage{multirow}
\usepackage{caption}
\usepackage{txfonts}
\usepackage{array}
\usepackage{threeparttable}    %主要需要载入这个包
\usepackage{booktabs}
\usepackage{longtable}
\usepackage{bm}
\usepackage{setspace}
\usepackage{tabularx}
\usepackage{booktabs}
\usepackage{chngpage}
\usepackage{float}
\usepackage{cases}
\makeatletter % changes the catcode of @ to 11
\newcommand{\linebreakand}{%
  \end{@IEEEauthorhalign}
  \hfill\mbox{}\par
  \mbox{}\hfill\begin{@IEEEauthorhalign}
}
\makeatother % changes the catcode of @ back to 12

\begin{document}
\title{Infinite forecast combinations based on Dirichlet process}
\author{\IEEEauthorblockN{Yinuo Ren}
\IEEEauthorblockA{\textit{Academy of Mathematics and Systems Science} \\
\textit{University of Chinese Academy of Sciences}\\
Beijing, China \\
renyinuo23@mails.ucas.ac.cn}
\and
\IEEEauthorblockN{Feng Li*}
\IEEEauthorblockA{\textit{School of Statistics and Mathematics} \\
\textit{Central University of Finance and Economics}\\
Beijing, China  (Corresponding author)\\
feng.li@cufe.edu.cn}
\linebreakand
\and
\IEEEauthorblockN{Yanfei Kang}
\IEEEauthorblockA{\textit{School of Economics and Management} \\
\textit{Beihang University}\\
Beijing, China\\
yanfeikang@buaa.edu.cn}
\and
\IEEEauthorblockN{Jue Wang}
\IEEEauthorblockA{\textit{Academy of Mathematics and Systems Science} \\
\textit{University of Chinese Academy of Sciences}\\
Beijing, China \\
wjue@amss.ac.cn}
\thanks{Presented in oral at the 2023 IEEE International Conference on Data Mining (ICDM): Artificial Intelligence for Time Series Analysis Workhsop (AI4TS).}
}

\maketitle
\begin{abstract}
Forecast combination integrates information from various sources by consolidating multiple forecast results from the target time series. Instead of the need to select a single optimal forecasting model, this paper introduces a deep learning ensemble forecasting model based on the Dirichlet process. Initially, the learning rate is sampled with three basis distributions as hyperparameters to convert the infinite mixture into a finite one. All checkpoints are collected to establish a deep learning sub-model pool, and weight adjustment and diversity strategies are developed during the combination process. The main advantage of this method is its ability to generate the required base learners through a single training process, utilizing the decaying strategy to tackle the challenge posed by the stochastic nature of gradient descent in determining the optimal learning rate. To ensure the method's generalizability and competitiveness, this paper conducts an empirical analysis using the weekly dataset from the M4 competition and explores sensitivity to the number of models to be combined. The results demonstrate that the ensemble model proposed offers substantial improvements in prediction accuracy and stability compared to a single benchmark model.
\end{abstract}
\section{Introduction}
In the early stages of time series forecasting development, the prevailing approach centered around constructing a single best-performing model as the ultimate solution. Whether it was a traditional model rooted in statistical theory or a result derived from a neural network within the machine learning paradigm, relying solely on a single forecast carried inherent limitations, including inadequate information extraction, the inability to capture intricate data features, and susceptibility to the influence of random factors.

The ''no free lunch" theorem also underscored the impossibility of a universal model that could be flawlessly applied to all datasets \cite{Wolpert585893}. Consequently, the concept of ensemble learning swiftly gained prominence in various prediction applications. Forecast combinations, in particular, began to exhibit their superiority when confronted with high parameter uncertainty. Simultaneously, the realm of probability forecasting drew upon Bayesian statistics principles, leveraging prior distributions to calculate posterior probabilities and enabling density comparisons and non-parametric regression techniques.
\subsection{Importance of forecast combination}
% In contemporary management decision-making practices, the predictive capability level has progressively emerged as a pivotal determinant in the effectiveness of strategies.
The exponential surge in data volume presents a formidable challenge in the realm of big data forecasting technology. Consequently, for the majority of time series data, although trend decomposition may precede model construction, it frequently proves arduous for a single model to successfully capture all requisite target features within a brief training period. To address this issue, the concept of forecast combinations emerged. This approach involves constructing multiple models within a single sequence of data and integrating the results obtained from various forecasting methods. Since its formal introduction in 1969, the concept of forecast combination has garnered attention. Furthermore, forecast combinations offer significant computational cost savings. In general, while the forecast accuracy of an individual weak learner may fall short of desired standards, it comes with reduced modeling complexity and substantially shorter training times. Theoretical evidence indicates that an ensemble of multiple weak learners outperforms the best individual learner when integrated into a unified framework \cite{Schapire1990TheSO}. Simultaneously, different learners can leverage diverse sources of information from the training data, leading to improved accuracy and enhanced generalization performance at a minimal cost\cite{WANG2022}.

Through the ensemble of multiple forecasting models, forecast combinations not only leverage the strengths of individual models but also mitigate their respective limitations. Furthermore, the pivotal role of trimming in the combination framework enhances its resistance to outliers and erratic data points \cite{wang2022look}.
% By incorporating forecast results generated through diverse methodologies or models, practitioners can significantly augment the precision and dependability of their predictions. This diversification ensures that the combined forecast encompasses a wider range of potential outcomes, ultimately enhancing its performance across various scenarios.
\subsection{Limitations of finite mixture model}
The number of models included in the combination process directly impacts its overall performance. In an ideal scenario, having an infinite number of models at the disposal would theoretically lead to significantly improved results. However, the practical limitations of data availability, computational resources, and time constraints make it impossible to work with an infinite number of models in reality.

Finite mixture model(FMM) assumes a fixed number of mixture components, and ascertaining the appropriate number of components can pose a significant challenge. Opting for too few components can lead to under-fitting, resulting in inadequate model performance while an excessive number of components may induce over-fitting, causing the model to fit noise in the data \cite{McLachlanPeel2000}. Secondly, FMM operates under the assumption that all components within the mixture adhere to the same parametric form, typically Gaussian distributions. However, this assumption may not hold when confronted with highly heterogeneous or multimodal data distributions, necessitating the exploration of alternative modeling approaches \cite{Fraley}. Furthermore, FMMs may encounter difficulties in handling high-dimensional data, thereby requiring the application of dimension reduction techniques or consideration of alternative modeling strategies to ensure tractability \cite{bishop2006pattern}.
\subsection{Ensemble Learning}
Ensemble learning methods harness the power of multiple individual models to improve predictions, often surpassing the performance of any single model. This approach has shown remarkable success across various domains, including classification, regression, anomaly detection, and recommendation systems. For instance, the N-BEATS model enhances time series forecasting by altering loss functions and prediction windows within a deep neural network structure \cite{oreshkin2019nbeats}. In contrast, the Heterogeneous Deep Forest model (Heter-DF) simultaneously selects decision trees, random forests, XGBoost, and LightGBM for class distribution estimation at each layer \cite{LIU2022}. To address volatility and variability, Zhang applied a dynamic error correction approach and employed a multi-objective optimization algorithm (NSGA-II) to achieve accurate and stable time series prediction \cite{ZHANG2021427}.

The Bias-Variance decomposition proposed highlights the balance between accuracy and diversity \cite{Geman6797087}. An increase in individual base model accuracy often leads to reduced ensemble diversity \cite{Lu}. To measure higher-order diversity, Brown introduced a decomposition formula based on information theory and entropy to explore correlations among three or more models in the overall ensemble. In neural networks, Yang and Wang used the weight vector of the last fully connected layer to calculate the Euclidean Distance between models, inversely proportional to similarity \cite{Yang9274468}. Effectively managing the trade-off between accuracy and diversity is a critical aspect of ensemble learning, determining the final prediction portfolio's performance.

Trimming algorithms play a crucial role in reducing the size of the ensemble model, simplifying synthesis complexity, and enhancing robustness \cite{Margineantu}. Typically, the number of models left at the end must be predetermined when trimming. A well-established ranking algorithm can also be used to select the top K models for the final prediction portfolio \cite{Zhang2019TwoStageBP}. Diversity Regularized Ensemble Pruning (DREP) is a pruning technique that applies regularization through the promotion of diversity, which is considered closely associated with the complexity of the hypothesis space, revealing the impact on the generalization performance of voting within the PAC learning framework \cite{li2012diversity}. The model pool trimming method defines the range of combined forecasts, while the share of a base learner's prediction results determines overall accuracy.
%diversity trimming
\section{Methodology}
\begin{figure}[htp]
    \centering
    \includegraphics[width=8cm]{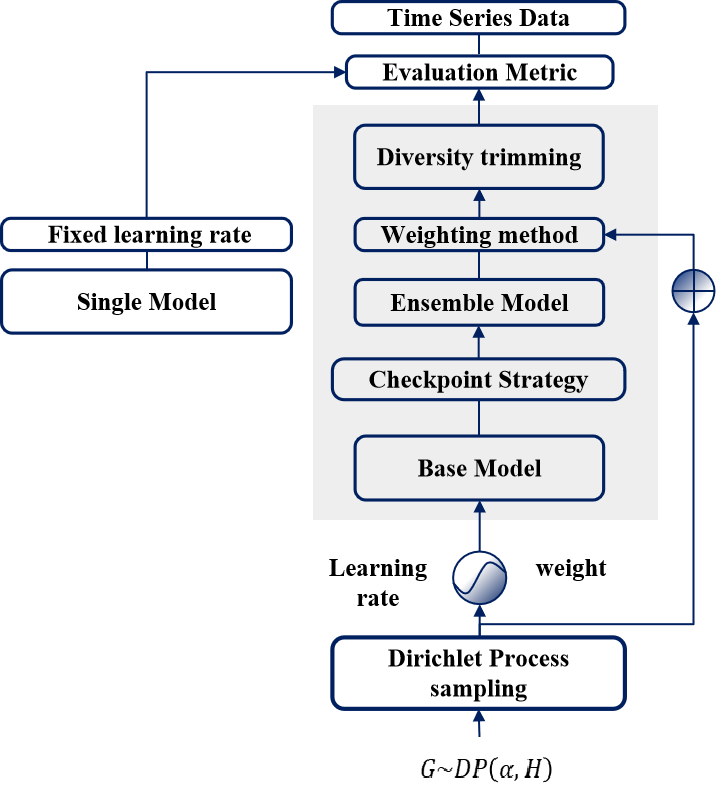}
    \caption{Ensemble model based on Dirichlet process.}
    \label{fig:structure}
\end{figure}

This section details the process of building an ensemble model based on the Dirichlet process and proposes a general framework in terms of the construction of the base learner and the setting of the ensemble strategy, in which the selection of neural networks and sampling parameters can be adjusted according to the actual needs.
Figure \ref{fig:structure} depicts the workflow for establishing the ensemble model. The sampling outcomes from the Dirichlet process determine the learning rate strategy for the base models and contribute to the weighting scheme of the ensemble model. Therefore, the selection of hyperparameters for the base distribution of the Dirichlet process is of utmost importance in this context.
Additionally, the ensemble framework undergoes experiments related to diversity trimming to further investigate the two key factors affecting ensemble performance: diversity and accuracy. In contrast, the single model, serving as the control group, simply maintains a fixed learning rate and utilizes its prediction errors as the baseline level in this experiment.
\subsection{Theoretic foundation}
\subsubsection{Combination with tuning parameters}
With the rapid expansion of computing resources and devices, the domain of forecasting has witnessed a surge in the adoption of complex machine learning algorithms (ML). These advanced machine learning methods have been widely applied in practical analyses and predictive competitions. They offer scalability and sophisticated adaptability, surpassing traditional statistical methods. However, this increased capability comes at the cost of reduced interpretability \cite{TEALAB2018334}. Within this array of machine learning approaches, neural networks distinguish themselves by their capacity to systematically analyze complex data through self-adjustment and autonomous learning. The multi-layer neural architecture equips them with exceptional information synthesis abilities, although it does pose challenges when it comes to parameter tuning. \cite{CRONE2011635}.

The choice of learning rate plays a pivotal role as it is a critical factor in gradient descent during the optimization of the loss function and determines the overall training time and the convergence to a local minimum. Due to the inherent randomness in the iterative process, even with identical models and datasets, results may exhibit variations, rendering it challenging to pinpoint the optimal learning rate \cite{bottou2018optimization}.

To address this issue, our approach in this study involves selecting a neural network as the single model and fixing the learning rate at a constant value for the majority of the experimental groups, to ensure the reliability of our research while maintaining control over variables. The base learner's framework remains consistent with the single model, except for the learning rate, forming the control group within the integrated neural network model.

\subsubsection{Dirichlet process}
Bayesian nonparametric Models are a class of statistical models that allow for a flexible approach when modeling complex data patterns, without the need for predefined parameters and can capture intricate data patterns but might need more data for good performance \cite{gershman2011tutorial}.
In this paper, we introduce an infinite mixture representation utilizing the Dirichlet process. The Dirichlet process (DP) is a stochastic process extensively employed in Bayesian nonparametric estimation. It generates samples from various distributions, earning it the moniker "a distribution of distributions" \cite{Teh}. The mathematical definition of the Dirichlet process is as follows:
\begin{align}
G \sim \text{DP}(\alpha, H).
\label{DP}
\end{align}
The DP allows us to generate probability distributions with infinite dimensions. As a positive scale parameter, $\alpha$ determines the dispersion of the base distribution $H$. When $\alpha = 0$, the sample taken is degenerated as one value; $\alpha \to \infty$, it can be equated to the base distribution $H$. Thus, each sampling of its samples is a distribution, hence it is also referred to as 'a distribution of distributions'.

% The sampling process of the Dirichlet process can be metaphorically likened to the process of breaking a unit-length stick into pieces. The specific procedure is as follows:
% \begin{itemize}
% \item \textbf{Step 1}: First, we take the initial sample $l_1 \sim H$. Then, the weight of this sample is set to $w_1$:
% \begin{align}
% \beta_1 \sim Beta(1,\alpha),\\
% w_1=\beta_1.
% \end{align}
% \item \textbf{Step 2}: Next, we sample the second sample $l_2 \sim H$ and the weight $w_2$:
% \begin{align}
% \beta_2 \sim Beta(1,\alpha),\\
% w_2=(1-w_1) \cdot \beta_2.
% \end{align}
% \item \textbf{Step 3}: Repeat the above steps continuing indefinitely. Eventually, we obtain a set of samples $l_{1},...,l_{n}$, all following $H$ distribution. And $w_1$ represents the weight they occupy \cite{Sudderth2006GraphicalMF}. The sampling process is illustrated as shown in Figure \ref{fig:stick-breaking}.
% \end{itemize}
% \begin{figure}[htp]
%     \centering
%     \includegraphics[width=6cm]{png/stick-breaking.png}
%     \caption{Dirichlet process sampling.}
%     \label{fig:stick-breaking}
% \end{figure}
\subsection{Ensemble prediction model based on Dirichlet process}
Schnaar et al. (1991) highlighted a two-stage process in the selection of base learners: firstly, selecting a suitable algorithm to create a model pool based on data features, and secondly, devising screening rules to enhance the final prediction \cite{SCHNAARS198671}.
According to Eq. \ref{DP}, it is evident that the samples drawn from the Dirichlet process are determined by the scaling parameter $\alpha$ and the base distribution $H$. Thus, the learning rate and the combination weight of the base learner $m_i$ are derived from Dirichlet process sampling and correspond to $\beta_i$ and $\pi_i$ parameters in the stick-breaking process $(i=1,2,...,p)$.

% While all base models share identical structures, the sole difference lies in the learning rates within the optimizer. Incorporating the learning rate decay algorithm alongside the Checkpoint strategies enables the training of all base learners in a single run, leading to substantial reductions in both computational costs and time \cite{Yang9274468}. The specific steps are as follows
% \begin{itemize}
% \item Sort the list of learning rates with a length of $p$ sampled from the Dirichlet process in descending order and substitute the sorted list into the optimizer.
% \item Specify the number of iterations for an individual model as $I$; therefore, after conducting $I$ iterations using the first learning rate, save the weight parameters, which will serve as the first base model denoted as $m_1$.
% \item Reduce the learning rate to the value at the second position in the sorted list, conducting $I$ iterations and save the weight parameters, designating the second base model $m_2$.
% \item Repeat the above steps to complete the training of all base models $m_i$ $(=1, 2, \ldots, p)$. The entire training process necessitates a total of $p \cdot I$ iterations.
% \end{itemize}
\subsubsection{Combination strategies}
Upon the conclusion of the training phase, the straightforward procedure entails loading the previously stored weight parameter files to reconstruct the neural network architectures of the base models in the prediction stage. Consequently, for each data sequence, an ensemble of $p$ prediction results is generated, and the ultimate prediction for that sequence is calculated by computing the weighted average value.

Indeed, as inferred from the previously outlined construction process, both the learning rate and combined weights are subject to sampling via the Dirichlet process. In other words, by specifying the number of base models $p$, we can ascertain a corresponding set of learning rates represented as $l$ and a combination weight vector denoted as $w$. As a result, the ensemble model $E$ can be represented as:
\begin{align}
    E = E(m, \alpha, H, p).
\end{align}
Under the condition of ensuring the validity and usability of the posterior computation, this paper provides a relevant theoretical research basis for the infinite extension of the hybrid model based on probability distribution. The contribution of this paper can be exhibited as follows:
\begin{itemize}
\item To tackle the problem of instability in results produced by neural networks trained with gradient descent, this paper generates base models by employing a segmented decay approach for learning rates, so that an ensemble model is constructed to mitigate uncertainties related to model selection and parameters.
\item This paper combines statistical distributions and stochastic processes to sample learning rates and weights through a Dirichlet mixture model and explores the application of threshold settings in the transformation of infinite into finite models.
\item Based on the concept of ensemble learning, this study examines the impact of base distribution diversity trimming and weight assignment strategies on prediction performance, and successfully harnesses the characteristics of multiple data patterns to optimize forecast combinations, achieving both robustness and accuracy simultaneously.
\end{itemize}
\section{Empirical Experiments}
\subsection{M4 Competition dataset}
\begin{figure}[htp]
    \centering
    \includegraphics[width=8cm]{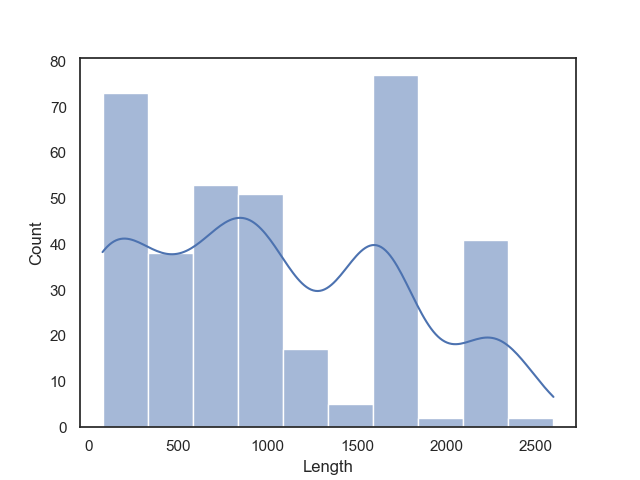}
    \caption{Histogram of weekly series length distribution.}
    \label{fig:weekly}
\end{figure}

Over the years, forecasting competitions have had a significant impact on the empirical domain, providing a solid foundation for assessing various inference methods and learning from experience to advance the practice of forecasting. The M4 competition dataset includes both high-frequency and low-frequency data and consists mostly of long sequences, providing more opportunities for complex methods that require extensive data for training. Therefore, we select it as the dataset for experimental research \cite{MAKRIDAKIS202054}.
% The M4 dataset comprises 100,000 time series meticulously gathered from the ForeDeCk database, curated by the National Technical University of Athens. These time series are drawn from a diverse range of public sources, including domains such as services, tourism, imports and exports, demographics, transportation, natural resources, and environmental data. The dataset spans six distinct data frequencies: annual, quarterly, monthly, weekly, daily, and hourly observations.
% The prediction horizon is determined based on the nature of the decisions most likely to be supported within companies or organizations for each data frequency. In this empirical analysis, we primarily utilize weekly data, comprising a total of 359 time series observations. These time series encompass various domains, with 112 originating from the microeconomic, 6 from the macroeconomic, 41 from the financial, 164 related to demographic data, and an additional 12 belonging to miscellaneous. Notably, all prediction horizons in the testing set are set to 13 time steps.
Figure \ref{fig:weekly} illustrates the histogram distribution of the length of the weekly series. The majority of series fall within the range of approximately 2000 weeks. The longest sequence spans up to 50 years, while the shortest sequence has a length of 276 weeks.
\subsection{Base models}
It is noteworthy that deep learning has become one of the most dynamic technologies across various research domains. Typically, machine learning tasks involve the stacking of multiple layers of neural networks and rely on stochastic optimization techniques to enhance generalization capabilities and improve predictive performance. In particular, Long Short-Term Memory Recurrent Neural Networks (LSTM), initially proposed by Hochreiter and Schmidhuber in 1997, have garnered significant attention in the field of time series analysis \cite{Hochreiter6795963}. LSTM differs fundamentally from traditional feedforward neural networks in its ability to establish correlations between past information and the current state, making it a sequence-based model. This implies that actions taken at $t-1$th step can impact decisions made at $t$th time steps.

% To establish temporal connections, the defining and maintaining of internal memory units are critical elements within the LSTM structure. These memory units play a key role in determining which elements in the state vector should be updated or retained, based on the output from the previous time step and the input from the current module. This process is accomplished by utilizing three Sigmoid functions as "soft" gates to determine which signals should be allowed to pass through:
% \begin{itemize}
%     \item \textbf{Input Gate}: Controls the information to retain in the internal state.

%     \item \textbf{Forget Gate}: Filters out information from the previous state that needs to be discarded.

%     \item \textbf{Output Gate}: Updates the internal state and determines which information should be passed on to the next module.
% \end{itemize}
The above process will continue in the $t+1$th step, repeating until the network construction is complete. Consequently, LSTM can adjust its parameters based on data characteristics to store and maintain information over different time steps, ultimately influencing the output of future modules.

\begin{figure*}[htp]
    \centering
    \includegraphics[width=14cm]{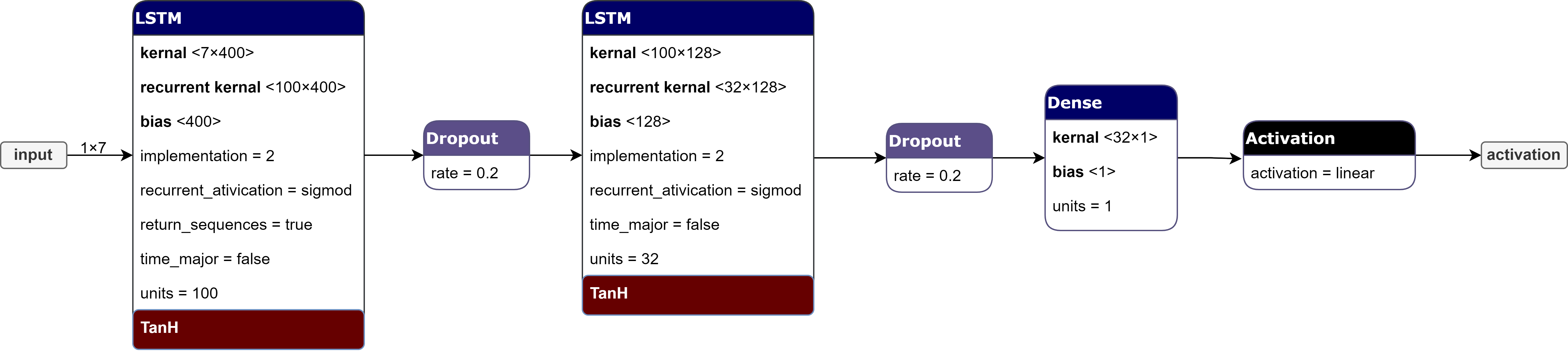}
    \caption{Neural network architecture of the base model.}
    \label{fig:LSTM}
\end{figure*}

Figure \ref{fig:LSTM} illustrates the schematic framework of the base model, where a lag of 7 steps is configured. It incorporates two LSTM modules and a Dropout layer, followed by a Dense layer and two activation functions to minimize the differences in weights and biases. To maintain consistency across other variables, a single model $S$ is designated as the experimental group, the structure of which mirrors that of base models, except for a fixed learning rate set at 0.001.
\subsection{Data Preprocessing}
Considering the sensitivity of LSTM layers to data scaling, it's necessary to merge data from train and test files and then scale it to [0, 1]. For univariate time series, the output value is the prediction at $t$th time step, and the corresponding input comes from $t-b$th to $t-1$th time step, where $b$ represents the lag steps. Thus, each row in the input matrix represents scaled features with a lag of $b$ time steps. The selection of $b$ typically depends on the practical periodicity of the sequence. For instance, in weekly data, $b$ might be set to values like 7 or 14.

Hence, the training dataset undergoes a format transformation, obtaining a $(T-b) \cdot b$ input matrix and a $(T-b)$ output variable vector, where $T$ represents the sequence length. In the prediction phase, the input takes the form of an $h-b$ data frame, with $h$ denoting the desired prediction horizon.
\subsection{Evaluation metric}
To illustrate the versatility of the ensemble framework in improving predictive performance it is essential to conduct error analysis across the entire weekly dataset $Y$ $(y_k, k=1,2, ..., N)$. When comparing the predictive accuracy of the ensemble model $E$ and the single model $S$, we define the prediction error of $E$ for the $k$th sequence $y_k$ as $Mstric(E, y_k)$ and $S$ as $Mstric(S, y_k)$. Subsequently, we calculate the average prediction errors for all sequences to derive the overall prediction errors on the weekly dataset $Y$:
\begin{align}
Metric(E, Y) = \frac{1}{N} \sum^N_{k=1} Metric(E, y_k),\\
Metric(S, Y) = \frac{1}{N} \sum^N_{k=1} Metric(S, y_k).
\end{align}
The evaluation of accuracy in time series forecasting typically involves the assessment of error deviation metrics, including root mean square error (RMSE) and mean absolute error (MAE):
\begin{align}
\mathrm{RMSE}=&\textstyle\sqrt{\left[\textstyle\sum_{t=1}^T(y_t-\hat{y}_t)^2\right]/T},\label{rmse}\\
\mathrm{MAE}=&\left(\textstyle\sum_{t=1}^T\left|y_t-\hat{y}_t\right| \right)/T.\label{mae}
\end{align}
Algorithm \ref{A1} reveals the specific implementation of the ensemble model $E$ based on the Dirichlet process.
\begin{algorithm}[htbp!]
\renewcommand{\algorithmicrequire}{\textbf{Input:}}
\renewcommand{\algorithmicensure}{\textbf{Output:}}
\caption{Ensemble prediction model based on Dirichlet process}\label{A1}
\begin{algorithmic}[1]
\REQUIRE Time series data $Y$ (Length $T$), Number of base models $p$, Hyperparameters $(\alpha H)$, Neural network $M$, Number of iterations $I$, Lag steps $b$.
\ENSURE Base model $m_i$, Ensemble model $E$.
\STATE Obtain a $(T-b) \cdot b$ training input matrix and a $(T-b)$ dependent variable vector after normalizing $Y$.
\STATE Derive a set of learning rates $l$ and combination weight vectors $w$ with the length of $p$ based on Eq. \ref{DP}.
\STATE Insert the descending learning rate list $l$ into the optimizer of $M$ and at every $I$ iterations save Checkpoint files using the decay algorithm, thereby completing the training of all base learners $m_i (i=1, 2,..., p)$.
\STATE Incorporate diversity trimming and weighting strategies to construct an ensemble prediction model and calculate prediction errors. The overall objective function is as: $arg min Metric(E, l, w, p)$.
\end{algorithmic}
\end{algorithm}
\section{Discussion}
This paper aims to tackle the challenge of determining the optimal learning rate through ensemble methods. Therefore, the sampled learning rate list should fluctuate around the experimental group (0.001). Additionally, considering diversity trimming, three base distributions are set as hyperparameters for the Dirichlet process: exponential distribution $EXP(0.001)$, Gaussian distribution $N(0.001,0.01)$ and beta distribution $Beta(1,1000)$. The corresponding ensemble models are denoted as $E_{exp}$, $E_N$ and $E_{beta}$ with scale parameters set to 1000 for each.
\subsection{Analysis of the impact of model number on prediction accuracy}
The Dirichlet process exhibits infinite possibilities, suggesting that the sample size $p$ can be extended to infinity in a single draw. However, to translate the infinite concept into a finite context for practical empirical analysis, this paper initiates experiments exploring how the number of models impacts the ensemble effect. This investigation serves as a crucial theoretical foundation for subsequently improving prediction accuracy.

\begin{figure*}[ht]
\centering
	\subfloat[MAE variation with the model number in $E$]{\includegraphics[width = 0.45\textwidth]{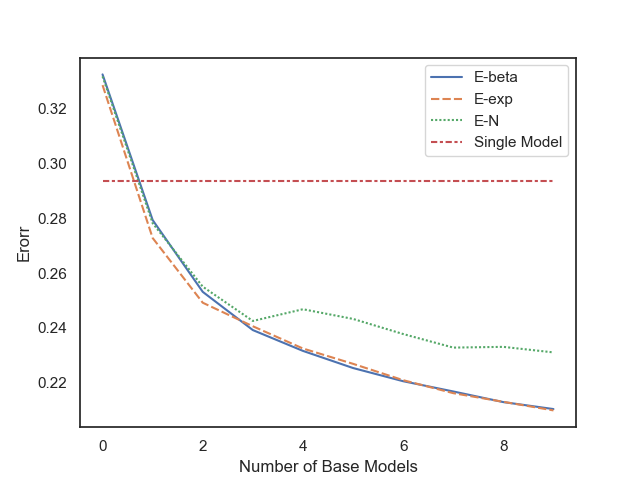}}
	\hfill
	\subfloat[RMSE variation with the model number in $E$]{\includegraphics[width = 0.45\textwidth]{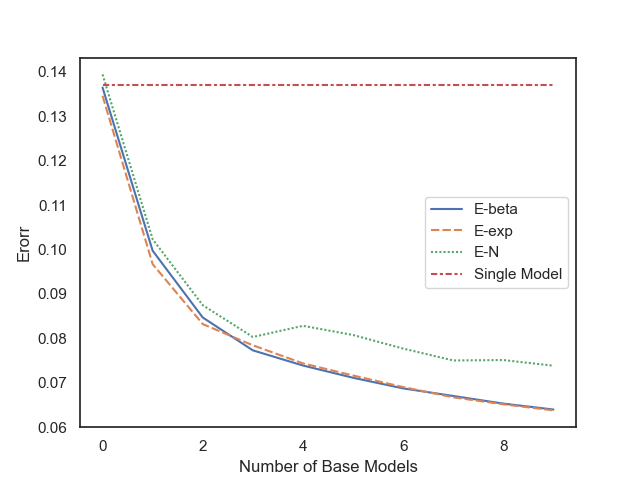}}
\caption{Line charts of prediction error varying with model number under three base distributions.}
\label{Number}
\end{figure*}

Fig. \ref{Number} depicts the changes in prediction error as a function of the number of models, considering three different base distributions.
Drawing insights from the above two figures, it becomes evident that when the number is 10 or less, the predictive performance of ensemble model $E$ significantly lags behind that of the single model $S$. However, as the number of base models increases, the forecast errors of the three ensemble models gradually diminish. Notably, the enhancement in performance is slightly more pronounced for $E_{exp}$ and $E_{beta}$ compared to $E_N$. The most substantial reduction in MAE and RMSE occurs in the range from 20 to 50 of $p$, resulting in a nearly 50\% decrease in errors compared to the single model $S$.
However, it is noteworthy that after reaching 60 models, the rate of improvement in forecast accuracy tends to slow down. This indicates that having a greater number of models does not necessarily lead to a proportionate increase in forecast accuracy. It further underscores the necessity of employing a model pool pruning strategy. Excessively large numbers of base models can result in wasteful pre-training computation costs and a significant increase in storage demands, all without achieving the desired prediction precision. Therefore, the selection of base learners remains a crucial component of constructing an ensemble framework.

\subsection{Analysis of the impact of model diversity on prediction accuracy}
Based on the foundational concept of the "variance-bias" decomposition formula \cite{Geman6797087}, it becomes apparent that accuracy and diversity are the two main factors that should be taken into account when designing an ensemble model. The theoretical framework introduced by Kang et al. (2020) through ambiguity decomposition serves as empirical evidence that a combination strategy exhibiting higher diversity yields a smaller overall error. In other words, a more diverse pool of base models directly contributes to an elevation in general prediction accuracy.

\begin{figure*}[htbp!]
\centering
	\subfloat[$p = 10$]{\includegraphics[width = 0.45\textwidth]{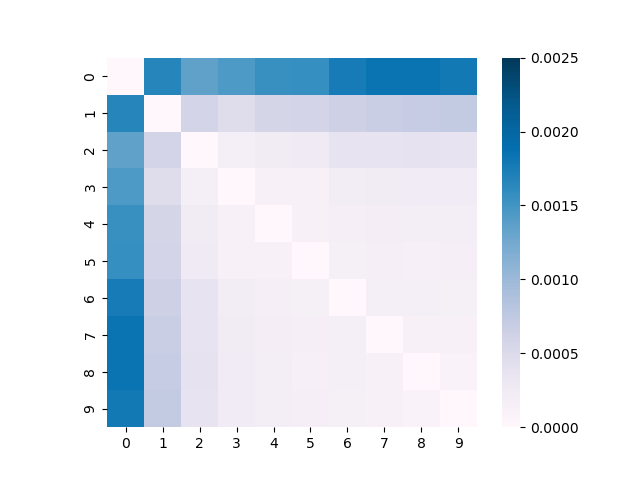}}
	\hfill
	\subfloat[$p = 20$]{\includegraphics[width = 0.45\textwidth]{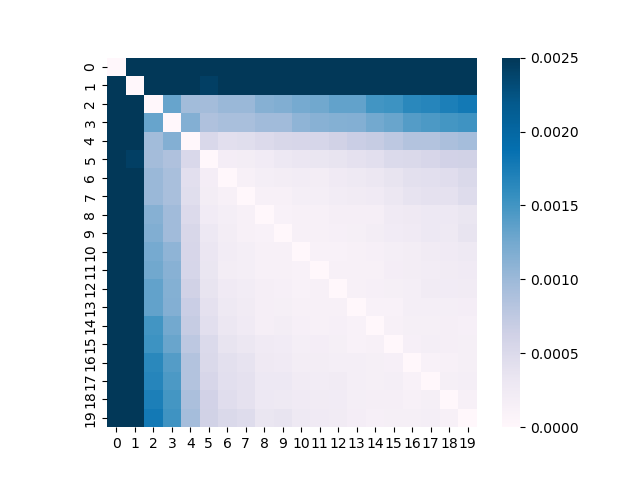}}
\caption{Ensemble model diversity heatmap.}
\label{Div}
\end{figure*}

To provide a more concrete understanding, we consider a specific time series denoted as $y_k$ and assign the $h$th step prediction generated by the $i$th base learner in the $k$th time series as $f_{ikh}$. Furthermore, $DIV_{ij}$ signifies the measure of diversity between the $i$th and $j$th base models within the prediction method pool. This diversity metric is rigorously defined as follows:
\begin{align}
    DIV_{i,j} = \frac{1}{H} \frac{1}{H} \sum^N_{k=1} \sum^H_{h=1} (f_{ikh} - f_{jkh})^2,
\end{align}
where $i,j = 1, 2,..., p$, and $h$ represents the prediction step with values spanning from 1 to $H$, while $N$ signifies the total number of series, $k = 1, 2,...,N$.  A higher value of $DIV_{i,j}$ indicates a greater degree of diversity within the ensemble model. Fig. \ref{Div} shows the diversity correlation matrices, and it can be seen that as the number of models increases, the diversity among the base learners steadily intensifies. This observation aligns with the earlier discussed trend of decreasing prediction error. Specifically, at a model number of 10, the lack of diversity renders $E$ unable to enhance prediction accuracy. However, when $p$ reaches 20, there is a significant order of magnitude increase in diversity, resulting in a remarkable reduction in prediction error.
The previous experiments primarily concentrated on establishing an ensemble model with the same base distribution for forecast combinations. It's worth noting that the homogeneity in base model training often results in a higher likelihood of positive correlation among outcomes. Hence, the augmentation of diversity plays a pivotal role in determining the predictive quality. Furthermore, it ensures that each base learner scrutinizes data features from a distinct perspective, thus contributing positively to the overall effectiveness.
In line with the principle of diversy trimming, this paper introduces a mixed strategy. Specifically, base models under each base distribution are selected and combined to create a more intricate ensemble model $E'$. $E'$ is then compared with the average prediction errors of $E_{exp}$, $E_{beta}$ and $E_N$.

\begin{figure}[htp]
    \centering
    \includegraphics[width=7cm]{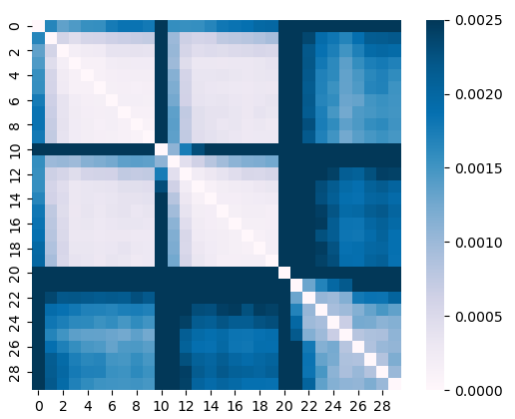}
    \caption{Mixed ensemble model diversity heatmap}
    \label{fig:Mix}
\end{figure}

Table \ref{t1} and Table \ref{t2} present the average performance metrics calculated over weekly data:
\begin{table*}[htbp!]
\scriptsize
\renewcommand\arraystretch{1.5}
\centering
\caption{MAE of $E'$ and the average MAE of $E_{exp}$, $E_{beta}$ and $E_N$.}
\label{t1}
\begin{threeparttable}
\setlength{\tabcolsep}{1.0mm}{
\begin{tabular}{lcccccccccc}
\toprule
Number of base models & 10 & 20 & 30 & 40 & 50 & 60 & 70 & 80 & 90 & 100 \\
\midrule
Mixed ensemble model $E'$ & 0.3308 & 0.2762 & 0.2520 & 0.2402 & 0.2368 & 0.2317 & 0.2262 & 0.2217 & 0.2197 & 0.2172  \\
Average MAE of $E_{exp}$, $E_{beta}$ and $E_N$ & 0.3310 & 0.2768 & 0.2524 & 0.2407 & 0.2369 & 0.2318 & 0.2264 & 0.2219 & 0.2196 & 0.2171 \\
\bottomrule
\end{tabular}}
\end{threeparttable}
\end{table*}
\begin{table*}[htbp!]
\scriptsize
\renewcommand\arraystretch{1.5}
\centering
\caption{RMSE of $E'$ and the average RMSE of $E_{exp}$, $E_{beta}$ and $E_N$.}
\label{t2}
\begin{threeparttable}
\setlength{\tabcolsep}{1.0mm}{
\begin{tabular}{lcccccccccc}
\toprule
Number of base models & 10 & 20 & 30 & 40 & 50 & 60 & 70 & 80 & 90 & 100 \\
\midrule
Mixed ensemble model $E'$ & 0.1365 & 0.0991 & 0.0847 & 0.0782 & 0.0765 & 0.0740 & 0.0712 & 0.0690 & 0.0679 & 0.0666  \\
Average RMSE of $E_{exp}$, $E_{beta}$ and $E_N$ & 0.1369 & 0.0996 & 0.0851 & 0.0787 & 0.0770 & 0.0745 & 0.0718 & 0.0696 & 0.0685 & 0.0672 \\
\bottomrule
\end{tabular}}
\end{threeparttable}
\end{table*}
It can be seen that $E'$ leverages three distinct base distributions to bolster the existing combination approach and obtain the expansion of the prediction pool's membership structure, significantly broadening the spectrum of available parameter settings and ultimately leading to substantial improvements in forecast performance when contrasted with a single-sampling approach. Fig. \ref{fig:Mix} complements these findings by presenting the diversity correlation matrix of $E'$, illustrating how the mixed strategy bolsters diversity within the ensemble framework.
Simultaneously, with the proliferation of models, the trajectory of prediction errors follows a pattern akin to our earlier analysis. However, it's essential to acknowledge that the associated computation costs rise exponentially. Hence, it becomes imperative to consider the limitations imposed by computational resources, ensuring a judicious balance between precision and diversity.
\subsection{Analysis of the impact of combination weights on prediction accuracy}
\begin{figure*}[htbp!]
\centering
	\subfloat[MAE]{\includegraphics[width = 0.45\textwidth]{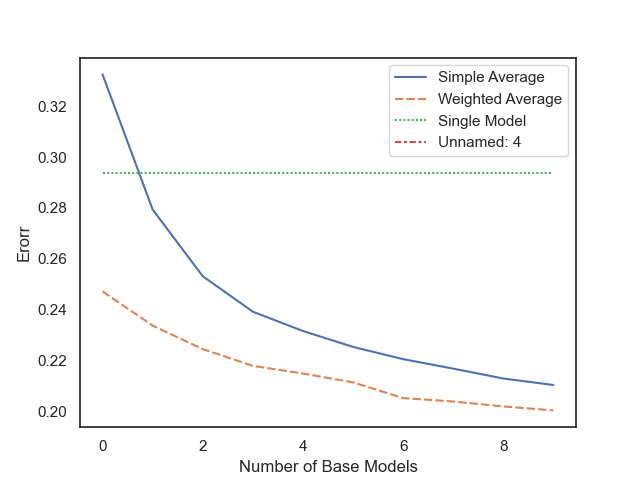}}
	\hfill
	\subfloat[RMSE]{\includegraphics[width = 0.45\textwidth]{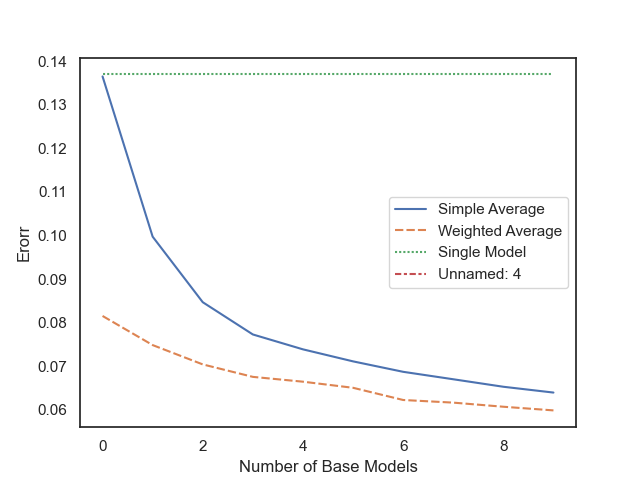}}
\caption{Forecast error variation with the model number of $E_{beta}$.}
\label{BT}
\end{figure*}

\begin{figure*}[htbp!]
\centering
	\subfloat[MAE]{\includegraphics[width = 0.45\textwidth]{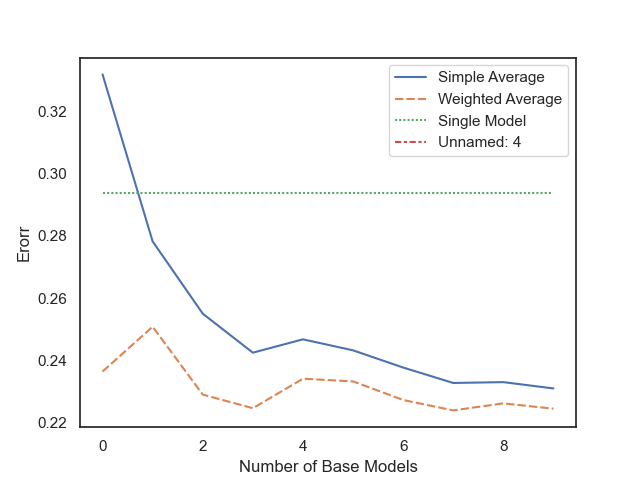}}
	\hfill
	\subfloat[RMSE]{\includegraphics[width = 0.45\textwidth]{png/EX_RMSE.png}}
\caption{Forecast error variation with the model number of $E_N$.}
\label{NM}
\end{figure*}

While in most cases, simple averaging serves as a fast and effective ensemble method, weighted averaging aims to go beyond uniform allocation by considering the varying importance of each base learner and determining their contribution to the terminal outcome. As mentioned above, the weights $w = (w_1, w_2,...,w_p)$ are derived from the sampling of the Dirichlet process. Therefore, the weight assigned to $m_i$ is defined as:
\begin{align}
    w_i' = \frac{w_i}{\sum^p_{i=1}w_i}.
\end{align}
Fig. \ref{EX}, \ref{BT} and \ref{NM} show the error analysis for $E_{exp}$, $E_{beta}$ and $E_N$ on weekly data using simple and weighted averages respectively.
To illustrate the universality of the results, Table \ref{t3} and \ref{t4} present a unified comparison by averaging the MAE and RMSE of the three ensemble frameworks under both simple averaging and weighted averaging methods:
\begin{figure*}[htbp!]
\centering
	\subfloat[MAE]{\includegraphics[width = 0.45\textwidth]{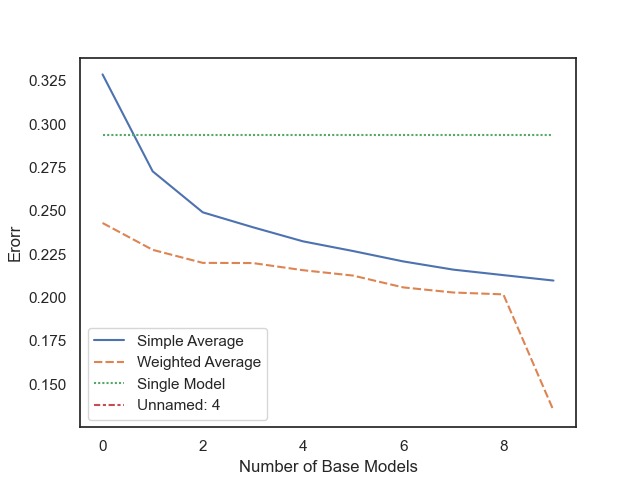}}
	\hfill
	\subfloat[RMSE]{\includegraphics[width = 0.45\textwidth]{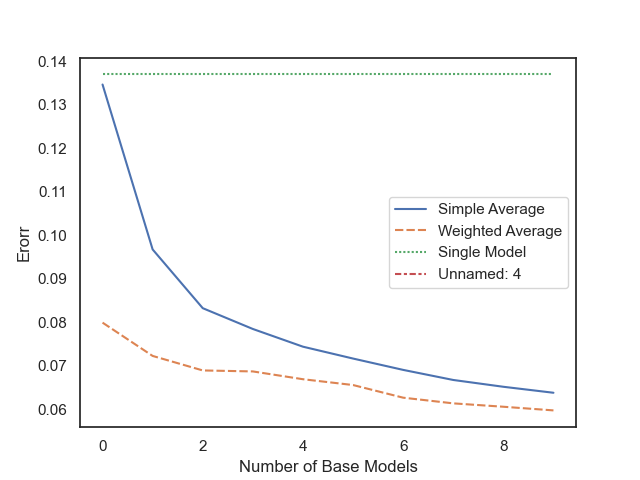}}
\caption{Forecast error variation with the model number of $E_{exp}$.}
\label{EX}
\end{figure*}
\begin{table*}[htbp!]
\scriptsize
\renewcommand\arraystretch{1.5}
\centering
\caption{Average MAE of $E_{exp}$, $E_{beta}$ and $E_N$.}
\label{t3}
\begin{threeparttable}
\setlength{\tabcolsep}{1.0mm}{
\begin{tabular}{lccccccccccc}
\toprule
Number of base models & 10 & 20 & 30 & 40 & 50 & 60 & 70 & 80 & 90 & 100 & Single model \\
\midrule
Simple average & 0.331 & 0.277 & 0.252	& 0.241 & 0.237 & 0.232 & 0.226 & 0.222 & 0.220 & 0.217& \multirow{2}{*}{0.294} \\
Weighted average & 0.137 & 0.100 & 0.085 & 0.079 & 0.077 & 0.074 & 0.072 & 0.070	& 0.069	& 0.067& ~ \\
\bottomrule
\end{tabular}}
\end{threeparttable}
\end{table*}
\begin{table*}[htbp!]
\scriptsize
\renewcommand\arraystretch{1.5}
\centering
\caption{Average RMSE of $E_{exp}$, $E_{beta}$ and $E_N$.}
\label{t4}
\begin{threeparttable}
\setlength{\tabcolsep}{1.0mm}{
\begin{tabular}{lccccccccccc}
\toprule
Number of base models & 10 & 20 & 30 & 40 & 50 & 60 & 70 & 80 & 90 & 100 & Single model \\
\midrule
Simple average & 0.137 & 0.100 & 0.085 & 0.079 & 0.077 & 0.074 & 0.072 & 0.070 & 0.069 & 0.067 & \multirow{2}{*}{0.137} \\
Weighted average & 0.083 & 0.075 & 0.071 & 0.069 & 0.070 & 0.069 & 0.065 & 0.064 & 0.064 & 0.063 & ~ \\
\bottomrule
\end{tabular}}
\end{threeparttable}
\end{table*}
As evident from the figures and tables above, the inclusion of weighted averaging in the ensemble method not only enhances forecast accuracy but also boosts the efficiency of ensemble learning compared to a single model. To contextualize this, when the number of models is small, the ensemble model might not outperform a single model. However, the weight strategy guarantees a reduction in error by nearly 50\% even when the number of models falls short, signifying that enhancements in weights can substantially curtail training time and storage requirements and ultimately yield low prediction error.
\section{Conclusion}
In this paper, we introduce an ensemble framework leveraging the Dirichlet process for time series forecast combinations. Our approach involves utilizing sampling to derive a set of learning rates and weights, which are then incorporated as new parameter inputs in the base models. This innovative strategy leads to a substantial enhancement in forecast accuracy. In contrast, the stochastic nature of gradient descent not only falls short of maintaining consistently high forecast accuracy but also introduces instability into the iterative backward process. Therefore, the primary objective of this study is to address the challenge faced by neural networks in determining the optimal learning rate.

% In response to this issue, we adopt a combination approach by configuring three base distributions (exponential distribution $EXP(0.001)$, Gaussian distribution $N(0.001,0.01)$ and beta distribution $Beta(1,1000)$) as hyperparameters for constructing the ensemble model. Since the base models have a shared structure, with differences only in their learning rate values, we utilize the Checkpoint strategy and a decay algorithm to save weight parameters at the same iteration intervals during a single training run. Consequently, during the forecasting phase, we can efficiently complete the training of all base models and directly load the saved files, significantly mitigating computational expenses and conserving memory space.

To assess the empirical effectiveness of the ensemble model, this study selects the weekly dataset from the M4 competition for analysis. The data preprocessing involves merging the training and test sets for each time series, followed by normalization. Subsequently, a lag step of 7 is applied to transform the data format into an input matrix. Finally, the average Root Mean Square Error (RMSE) and Mean Absolute Error (MAE) are calculated across all series data, enabling a comparison between the prediction performance of the ensemble model and the single model.

The findings reveal the following insights: (1) The ensemble model with equal weights significantly reduces forecast error, with a reduction of nearly 50\% observed when the number of models reaches 20 and above; However, the rate of error reduction tends to stabilize after reaching 60 models. (2) Utilizing the weighted averaging method proves to be more efficient in enhancing forecast accuracy; Remarkably, even with only 10 base models, the ensemble model demonstrates a considerable reduction in forecast error. (3) Embracing the concept of diversity, the implementation of a mixed strategy contributes to the construction of a more intricate framework; Combining three smaller ensemble models, to some extent, positively enhances the quality of combination forecasts.

By the fundamental ``variance-bias" decomposition principle \cite{Geman6797087}, addressing the trade-off between diversity and accuracy has consistently presented a formidable challenge for ensemble learning. This challenge arises from two critical factors: Firstly, the cornerstone of combination forecasting is the formation of model pools. When training data exhibits substantial similarity, it naturally leads to homogeneity and strong correlation among base learners. The key issue here is to extract data features from diverse perspectives, highlighting the essential role of diversity. Secondly, as the accuracy of an ensemble model improves, the required training time and costs grow exponentially.

% In this paper, leveraging the properties of the Dirichlet process, we narrow down the scope of the prediction pool by investigating the impact of varying model numbers. This transformation converts an infinite mixture into a finite one. Building upon this foundation, we introduce weight adjustment and diversity trimming strategies, collectively yielding a highly efficient and accurate algorithm tailored applied to time series forecast.

% This study has certain limitations that warrant acknowledgment. Due to constraints related to time and computational resources, we are unable to validate the algorithm's generalization capabilities on datasets with different frequencies from the M4 dataset or real-world data. Additionally, there remains room for further exploration in determining the optimal number of models, aiming to minimize computational resource wastage.
In our future research efforts, we intend to In future work, the article will select other ensemble models as benchmark models for comparison and implement a data-driven approach to establish a sampling threshold.
% \section*{Acknowledgments}

\section*{Acknowledgments}
We thank the reviewers for helpful comments that improve the contents of this paper. This research was supported by the National Social Science Fund of China (22BTJ028, 22VRC056), the National Natural Science Fund of China (72271229, 71771208).
\bibliographystyle{IEEEtran}
\bibliography{mybibefile}
\end{document}